\documentclass[10pt,twocolumn,twoside]{IEEEtran} 
\usepackage{times}
\usepackage{epsfig}
\usepackage{graphicx}
\graphicspath{{figures/}}
\usepackage{amsmath}
\usepackage{amssymb}

\usepackage{multirow}
\usepackage{array}
\usepackage{rotating}

\usepackage{xcolor}
\usepackage{url}
\usepackage{float}
\usepackage{subfigure}
\usepackage{booktabs}
\usepackage{bbding}
\usepackage{pifont}

\usepackage[english]{babel}

\ifCLASSINFOpdf
\else
\fi

\begin{document}
%

\title{Connecting Image Denoising and High-Level Vision Tasks via Deep Learning}

%
%

\author{Ding~Liu,~\IEEEmembership{Student Member,~IEEE,}
        Bihan~Wen,~\IEEEmembership{Student Member,~IEEE,}
        Jianbo~Jiao,~\IEEEmembership{Student Member,~IEEE,}
        Xianming~Liu,
        Zhangyang~Wang,~\IEEEmembership{Member,~IEEE,}
        and~Thomas~S.~Huang,~\IEEEmembership{Life~Fellow,~IEEE}
\thanks{D. Liu, J. Jiao and T. S. Huang are with the Beckman Institute, Univerisity of Illinois at Urbana-Champaign, Urbana,
IL, 61801 USA e-mail: (dingliu2@illinois.edu; jianboj@illinois.edu; t-huang1@illinois.edu).}
\thanks{B. Wen is with the Department of Electrical and Computer Engineering, Univerisity of Illinois at Urbana-Champaign, Urbana,
IL, 61801 USA e-mail: (bwen3@illinois.edu).}
\thanks{X. Liu is with Facebook Inc., San Francisco, CA 94025 USA (e-mail:	xmliu@fb.com).}
\thanks{Z. Wang is with the Department of Computer Science and Engineering, Texas A\&M University, TX 77843 USA (e-mail: atlaswang@tamu.edu).}
}

\maketitle

\begin{abstract}

Image denoising and high-level vision tasks are usually handled independently in the conventional practice of computer vision, and their connection is fragile.
In this paper, we cope with the two jointly and explore the mutual influence between them with the focus on two questions, namely (1) how image denoising can help improving high-level vision tasks, and (2) how the semantic information from high-level vision tasks can be used to guide image denoising. 
First for image denoising we propose a convolutional neural network in which convolutions are conducted in various spatial resolutions via downsampling and upsampling operations in order to fuse and exploit contextual information on different scales.
Second we propose a deep neural network solution that cascades two modules for image denoising and various high-level tasks, respectively, and use the joint loss for updating only the denoising network via back-propagation. 
We experimentally show that on one hand, the proposed denoiser has the generality to overcome the performance degradation of different high-level vision tasks.
On the other hand, with the guidance of high-level vision information, the denoising network produces more visually appealing results. 
Extensive experiments 
demonstrate the benefit of exploiting image semantics simultaneously for image denoising and high-level vision tasks via deep learning.
The code is available online: \url{https://github.com/Ding-Liu/DeepDenoising}

\end{abstract}


%
\IEEEpeerreviewmaketitle


\section{Introduction}


Conventionally, low-level image processing problems, such as image restoration and enhancement, and high-level vision tasks are handled separately by different approaches in computer vision.
In this work, we connect them by showing the mutual influence between the two, i.e., visual perception and semantics, and propose a new perspective for solving both the low-level image processing and high-level computer vision problems in a single unified framework. 

Image denoising, as one representative of low-level image processing problems, aims to recover the underlying image signal from its noisy measurement.
Conventional image denoising approaches exploit either local or non-local image characteristics \cite{Aharon2006,dabov2007image,Mairal2009,dong2013nonlocally,gu2014weighted,xu2015patch}.
In recent years, we have witnessed the revival of deep learning in computer vision and deep neural networks have been developed for image denoising with notable performance improvement~\cite{vincent2008extracting,burger2012image,mao2016image,chen2017trainable,zhang2017beyond}.
In this work, we propose a new convolutional neural network for image denoising.
Inspired by U-Net~\cite{ronneberger2015u}, we conduct convolutions in different spatial scales via downsampling and upsampling operations before the resulting features are fused together, so that the kernels have a larger receptive field after all feature contraction. 
Downsampling operations also help to reduce the computation cost due to the reduction of feature map sizes.
We conduct extensive experiments and show that our proposed denoising network achieves the state-of-the-art performance across benchmark datasets. 

Many popular image denoising approaches 
only minimizes the mean square error (MSE) between the reconstructed and clean images
whereas important image details are sometimes lost which results in image quality degradation; e.g., oversmoothing artifacts in some texture-rich regions are commonly observed in the denoised result from conventional methods. 
Meanwhile, the semantical information of the image is usually ignored during denoising. 
To overcome this drawback, we propose to cascade the network models for image denoising and a high-level vision task, respectively.
During training, we jointly minimize the image reconstruction loss, the perceptual loss over the feature domain, 
and the high-level vision loss.
With the guidance of image semantic information, the denoising network is able to further improve visual quality and generate more visually appealing results, which demonstrates the importance of semantic information for image denoising.

When high-level vision tasks are conducted on noisy data, 
image restoration is typically applied as an independent preprocessing step, which might be suboptimal for the ultimate goal~\cite{wang2016studying,wu2017relation,liu2017enhance}.
Recent research reveals that neural networks trained for image classification can be easily fooled by small noise perturbation or other artificial patterns~\cite{szegedy2013intriguing,nguyen2015deep}.
However, to the best of our knowledge, the problem of how low-level image processing could affect high-level semantical tasks is still not thoroughly studied~\cite{wang2016studying,wu2017relation}. 
In this work, we observe an intriguing result that applying a conventional denoising method such as CBM3D~\cite{dabov2007color} over noisy images may introduce lethal artifacts,
which results in unsatisfactory results of image classification and semantic segmentation.
Therefore, we aim to produce an application-driven denoiser which is capable of simultaneously removing noise and preserving semantic-aware details for the high-level vision tasks.
To this end, in our proposed cascaded architecture
we keep the high-level vision network untouched but only use the gradients of the joint loss to update the denoising network, in order to make sure the denoising network has good generalization about other high-level vision tasks. This also ensures that the denoising module removes noises while preserving the important semantic details to produce correct high-level vision task results. 

We systematically investigate the mutual influence between the low-level image denoising and high-level vision networks using our proposed architecture via numerous experiments.
We show that the cascaded network trained with the joint loss not only boosts the perceptual quality of denoised images via image semantic guidance, but also substantially improves the accuracy of high-level vision tasks.
Moreover, our proposed training strategy makes the trained denoising network generalizable to different high-level vision tasks. 
In other words, our denoising module trained for one high-level vision task can be directly plugged into other high-level tasks without fine-tuning either module,
which facilitates the training effort when applied to various high-level vision tasks and keeps the high-level vision networks performing consistently for noisy and noiseless images.

In short, the main contributions of this paper are as follows:
\begin{itemize}
    \item To the best of our knowledge, this is the first attempt to investigate the benefit of exploiting image semantics simultaneously for image denoising and high-level vision tasks under a unified deep learning framework.
	\item We demonstrate that high-level semantics can be used for image denoising to generate visually appealing results in a deep learning fashion.
	\item Our proposed training strategy enables the robustness of the trained denoising network to various high-level vision tasks so that it can directly work with various high-level vision networks without any fine-tuning.
\end{itemize}

This paper is built upon our previous work \cite{liu2018image} with several notable improvements.
First, compared with \cite{liu2018image} we additionally incorporate the perceptual loss over the feature domain in our combined loss in order to capture more semantic information from multilevel feature domains.
Second, we provide a detailed ablation study of the denoising network architecture, and we show an analysis of the gradient visualization of the cascade architecture, in order to further investigate how high-level image semantics guides low-level image denoising.
Third, we conduct a subjective evaluation on several recent image denoising methods to thoroughly measure the visual quality of denoised results.
Finally, we present a more comprehensive experiment section including comparisons of denoising results over more benchmark datasets, and we extend our proposed denoising method to real noise removal.

The remainder of this paper is organized as follows. Section~\ref{sec:related}
reviews existing image denoising methods in the literature. Section~\ref{sec:method}
presents our image denoising network and the proposed cascaded architecture. Section~\ref{sec:exp} provides the experimental results. Section~\ref{sec:conc} concludes this paper.

\section{Related Work}
\label{sec:related}

Denoising is the task of estimating the high-quality signal from its noisy measurements. 
Classical image denoising methods take advantage of local or non-local structures presented in the image explicitly.
Natural images are well-known to be patch-wise sparse or compressible in transform domain, or over certain dictionary. 
Prior works \cite{Aharon2006,zoran2011learning,wen2015octobos} exploit such local structures and reduce noise by coefficient shrinkage for image restoration. 
The later approaches, including SSC \cite{Mairal2009}, CSR \cite{Dong2011}, NCSR \cite{dong2013nonlocally}, GSR \cite{zhang2014group}, WNNM \cite{gu2014weighted}, PCLR \cite{chen2015external}, PGPD \cite{xu2015patch}, STROLLR \cite{wen2017strollr}, as well as BM3D \cite{dabov2007image} and its extension for color images CBM3D~\cite{dabov2007color} -- group similar patches within the image globally via block matching or clustering, and impose non-local structural priors on these groups, which usually lead to state-of-the-art image denoising performance.

More recently, the popular deep neural network techniques have been applied to low-level vision tasks.
Specifically, a number of deep learning models have been developed for image denoising~\cite{vincent2008extracting,jain2009natural,burger2012image,xie2012image,mao2016image,kim2017deeply},
which can be classified into two categories:    multilayer perception (MLP) based models and convolutional neural network (CNN) based models. 
Early MLP based models for image denoising include the stacked denoising autoencoder~\cite{vincent2008extracting},
which is an extension of the stacked autoencoder originally designed for unsupervised feature learning.
A denoising autoencoder is proposed for both image denoising and blind image inpainting by Xie et al.\ \cite{xie2012image}.
Burger et al.\ \cite{burger2012image} introduce a plain MLP and thoroughly compare its denoising performance with 
BM3D~\cite{dabov2007image} in different experimental settings.
CNNs are first utilized for image denoising by Jain and Seung~\cite{jain2009natural}.
Recent works \cite{schmidt2014shrinkage,chen2017trainable} attempt to unfold the iterative algorithms, and construct a cascaded convolutional filtering architecture for image denoising.
Very deep networks are developed with skip connections for image restoration~\cite{mao2016image,zhang2017beyond}, and achieve superior performance over other recent methods.
Dilated convolutions are adopted to learn the residual image for denoising in \cite{zhang2017learning}.
A CNN is designed to implicitly
learn the regularizer of the alternating minimization  algorithm for regularization-based image restoration methods in \cite{kim2017deeply}.
The formatted
residual information between the noiseless image and its denoised version is learned via a second network in \cite{jiao2017formresnet}. 
The non-local self-similarity property of natural
images are exploited with CNNs in \cite{lefkimmiatis2017non,lefkimmiatis2018universal}.
FFDNet is proposed in \cite{zhang2018ffdnet} to work  on downsampled images in order to reduce the inference time.
Non-local self-similarity of images is exploited and incorporated into a recurrent neural network in \cite{liu2018non}.

\begin{figure*}[th]
	\centering
	\includegraphics[width=\linewidth]{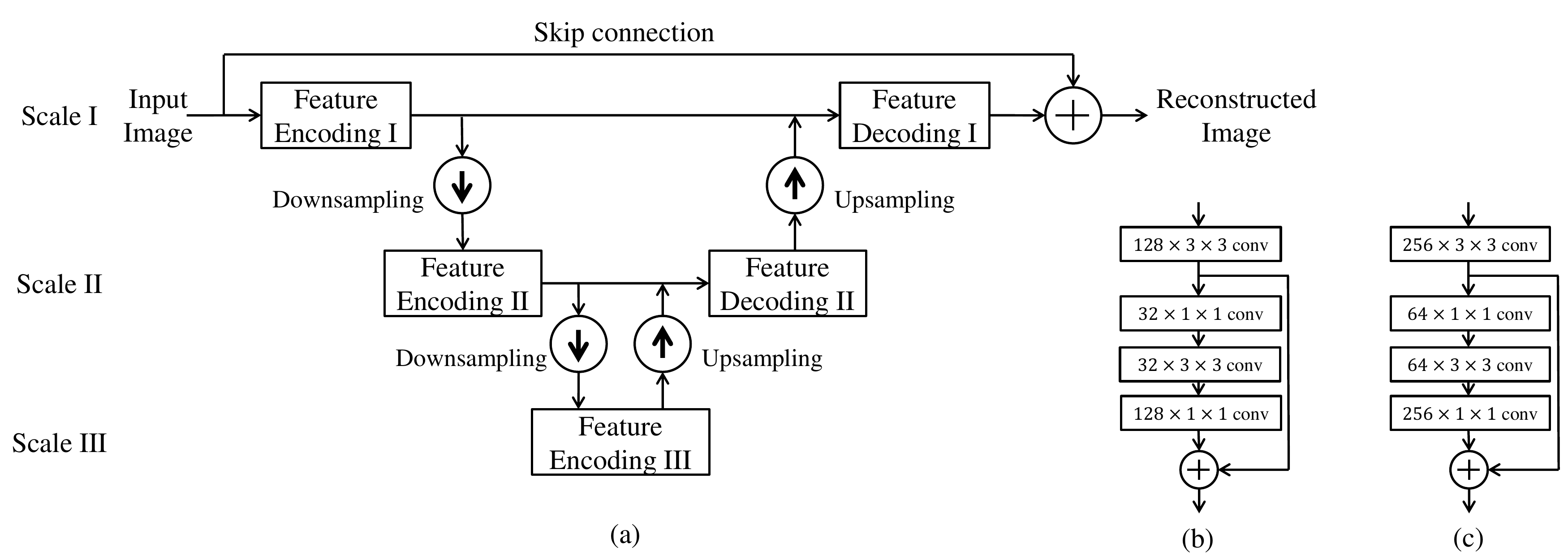}
	\caption{(a) Overview of our proposed denoising network. (b) Architecture of the feature encoding module. (c) Architecture of the feature decoding module.}
	\label{fig:denoising}
\end{figure*}

\section{Method}
\label{sec:method}

In this section, we first introduce the denoising network utilized in our framework, 
and then explain our framework that cascades this image denoising module and the module for high-level vision task with its training strategy and joint loss in detail.

\subsection{Denoising Network}
\label{lab:denoising}


We propose a convolutional neural network for image denoising, which takes a noisy image as input and outputs the reconstructed image. 
This network conducts feature contraction and expansion through downsampling and upsampling operations, respectively. 
Each pair of downsampling and upsampling operations brings the feature representation into a new spatial scale, so that the whole network can process information on different scales.

Specifically, on each scale, the input is encoded  after downsampling the features from the previous scale.
After feature encoding and decoding possibly with features on the next scale, the output is upsampled and fused with the feature on the previous scale.
Such pairs of downsampling and upsampling steps can be nested to build deeper networks with more spatial scales of feature representation, which generally leads to better restoration performance.
Considering the trade-off between computation cost and restoration accuracy, we choose three scales for the denoising network in our experiments, while this framework can be easily extended for more scales.
The ablation study of the denoising network architecture can be found in Section~\ref{sec:ablation}.

These operations together are designed to learn the residual between the input and the target output and recover as many details as possible, 
so we use a long-distance skip connection to sum the output of these operations and the input image, in order to generate the reconstructed image.
The overview is in Figure \ref{fig:denoising} (a).
Each module in this network will be elaborated on as follows.

\textbf{Feature Encoding}: We design one feature encoding module on each scale, which is one convolutional layer plus one residual block as in \cite{he2016deep}. 
The architecture is displayed in Figure~\ref{fig:denoising} (b). 
Note that each convolutional layer is immediately followed by spatial batch normalization and a ReLU neuron. From top to down, the four convolutional layers have 128, 32, 32 and 128 kernels in size of $3 \times 3, 1 \times 1, 3 \times 3$ and $1 \times 1$, respectively.
The output of the first convolutional layer is passed through a skip connection for element-wise sum with the output of the last convolutional layer.


\begin{figure*}[t]
	\center
	\includegraphics[width=\linewidth]{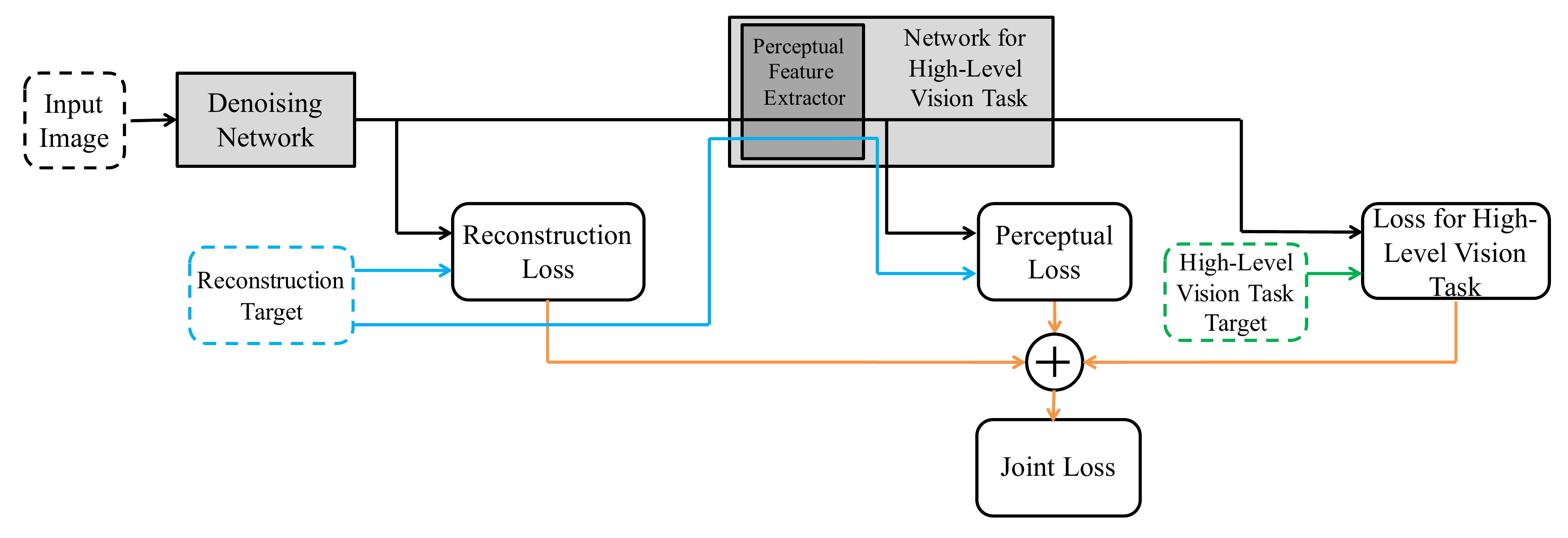}
	\caption{Overview of our proposed cascaded network.}
	\label{fig:system}
\end{figure*}

\textbf{Feature Decoding}: The feature decoding module is designed for fusing information from two adjacent scales.
Two fusion schemes are tested: (1) concatenation of features on these two scales; (2) element-wise sum of them. 
Both schemes obtain similar denoising performance.
Thus we choose the first scheme to accommodate feature representations of different channel numbers from two scales.
We use a similar architecture as the feature encoding module except that the numbers of kernels in the four convolutional layers are 256, 64, 64 and 256. 
Its architecture is in Figure~\ref{fig:denoising}(c).

\textbf{Feature Downsampling \& Upsampling}: Downsampling operations are adopted multiple times to progressively increase the receptive field of the following convolution kernels and to reduce the computation cost by decreasing the feature map size.
The larger receptive field enables the kernels to incorporate 
larger spatial context for denoising.
We use 2 as both the downsampling factor and the upsampling factor, 
and try two schemes for downsampling in the experiments: (1) max pooling with stride of 2; (2) conducting convolutions with stride of 2. Both schemes achieve similar denoising performance in practice, so we use the second scheme in the rest of the experiments for computation efficiency. 
Upsampling operations are implemented by deconvolution with $4 \times 4$ kernels, which aim to expand the feature map to the same spatial size as the previous scale.

Since all the operations in our proposed denoising network are spatially invariant, it has the merit of handling input images of arbitrary size.


\subsection{When Image Denoising Meets High-Level Vision Tasks}

We propose a robust deep architecture processing a noisy image input, via cascading a  network for denoising and the other for high-level vision task, aiming to simultaneously: 
\begin{enumerate}
	\item reconstruct visually pleasing results guided by the high-level vision information, as the output of the denoising network;
	\item attain sufficiently good accuracy across various high-level vision tasks, when trained for only one high-level vision task.
\end{enumerate}
The overview of the proposed cascaded network is displayed in Figure \ref{fig:system}.
Specifically, given a noisy input image, the denoising network is first applied, and the denoised result is then fed into the following network for high-level vision task, which generates the high-level vision task output.

\textbf{Training Strategy}:
First we initialize the network for high-level vision task from a network that is well-trained in the noiseless setting.
We train the cascade of two networks in an end-to-end manner while
fixing the weights in the network for high-level vision task.
Only the weights in the denoising network are updated by the error back-propagated from the following network for high-level vision task,
which is similar to minimizing the perceptual loss for image super-resolution~\cite{johnson2016perceptual}.
The reason for adopting such a training strategy is to make the trained denoising network robust enough without losing the generality for various high-level vision tasks. 
More specifically, our denoising module trained for one high-level vision task can be directly plugged into networks for other high-level tasks without fine-tuning either the denoiser or the high-level network.
Our approach not only facilitates the training effort when applying the denoiser to different high-level tasks while keeping the high-level vision network performing consistently for noisy and noise-free images, but also enables the denoising network to produce high-quality perceptual and semantically faithful results.

\textbf{Loss}:
Recent works \cite{gatys2015texture,gatys2016image} discover that multi-level feature statistics extracted from a trained CNN notably represent the visual semantics in various scales.
We aim to use this property to enhance the quality of the denoised results from our framework.
There are three loss components adopted in our framework, and we describe them sequentially as follows.

The reconstruction loss of the denoising network is the mean squared error (MSE) between the denoising network output and the noiseless image, which can be represented as
\begin{equation}
L_D(x, \tilde{x}) = \frac{1}{HW}\sum^{H}_{i=1}\sum^{W}_{j=1}(F_D(x)_{i,j} - \tilde{x}_{i,j})^2 \,,
\label{eqn:loss_denoising}
\end{equation}
\noindent
where $x$ is the noisy input image and $\tilde{x}$ is the noiseless image. $F_D$ denotes the denoising network. $x$, $\tilde{x}$ and $F_D(x)$ are of size $H \times W$. $\tilde{x}_{i,j}$ indicates the pixel in the $i$-th column and the $j$-th row of $\tilde{x}$.

In addition to the reconstruction loss calculated on the image domain, we define the perceptual loss as the euclidean distance between the feature
representations of a reconstructed image $F_D(x)$ and the
reference image $\tilde{x}$~\cite{johnson2016perceptual}. Here we choose the perceptual feature extractor $\phi(\cdot)$ to be the first several layers of the high-level vision network. Specifically, when we adopt VGG-based networks for image classification and semantic segmentation in our experiments, we use the feature $relu3\_3$ as the perceptual feature representation. The perceptual loss can be represented as
\begin{equation}
L_P(F_D(x), \tilde{x}) = \frac{1}{H_\phi W_\phi}\sum^{H_\phi}_{i=1}\sum^{W_\phi}_{j=1}(\phi(F_D(x))_{i,j} - \phi(\tilde{x})_{i,j})^2 \,,
\label{eqn:loss_perceptual}
\end{equation}
\noindent
where the perceptual feature representations of $\phi(\tilde{x})$ and $\phi(F_D(x))$ are of size $H_\phi \times W_\phi$.

The losses of the classification task and the segmentation task both are the cross-entropy loss between the predicted label and the ground truth label.
The joint loss is defined as the weighted sum of the reconstruction loss, the perceptual loss and the loss for high-level vision task, which can be represented as
\begin{equation}
L(x, \tilde{x}, y) = L_D(x, \tilde{x}) + \lambda_P L_P(F_D(x), \tilde{x}) + \lambda_H L_H(F_D(x), y),
\label{eqn:loss}
\end{equation} 
\noindent
where $y$ is the ground truth label of high-level vision task. 
$L_H(F_D(x), y)$  represents the loss of the high-level vision network with $F_D(x)$ as the input, and $L$ is the joint loss, as illustrated in Figure~\ref{fig:system}. $\lambda_P$ and $\lambda_H$ are the weights for balancing the losses $L_D$, $L_P$ and $L_H$.

\section{Experiments}
\label{sec:exp}

In this section, we first show the experiment results of our proposed denoising network, and then discuss its relation with high-level vision networks. Finally, we extend our denoising network to the case of real noise removal.

\begin{table*}[t]
	\centering
	\caption{Different network designs and their corresponding denoising performances. We finally choose the architecture in bold for our denoising network.}
	\vspace{0.05in}	
		\begin{tabular}{|c|c|c|c|c|c|}
			\toprule
			Scale No.     & Skip connection  & Encoding module &    Decoding module  & Parameters & PSNR   \\ 
			\midrule
			2  &\Checkmark& (128, 32, 32, 128)   &  (128, 32, 32, 128)   &   384k    & 30.41 \\			
			2  &\ding{55}&  (128, 32, 32, 128)  &  (256, 64, 64, 256)   &   637k    & 30.39 \\ 	
			2  &\Checkmark& (128, 32, 32, 128)   &  (256, 64, 64, 256)   &   637k    & 30.54 \\
			2  &\Checkmark& (256, 64, 64, 256)   &  (256, 64, 64, 256)   &   1780k    & 30.70 \\			
			\midrule
			3  &\Checkmark& (128, 32, 32, 128) & (128, 32, 32, 128) & 592k & 30.64 \\								
			3  &\ding{55}&  (128, 32, 32, 128)  &  (256, 64, 64, 256)   &   1,026k & 30.52 \\ 				
			\textbf{3}  &\textbf{\Checkmark}& \textbf{(128, 32, 32, 128)} & \textbf{(256, 64, 64, 256)} &   \textbf{1,026k} & \textbf{30.83} \\					
			3  &\Checkmark& (256, 64, 64, 256) & (256, 64, 64, 256) &   2,198k    & 30.87 \\					
			\midrule			
			4  &\Checkmark& (128, 32, 32, 128)   &  (128, 32, 32, 128)   &   970k    & 30.76 \\			
			4  &\ding{55}&  (128, 32, 32, 128)  &  (256, 64, 64, 256)   &   1,488k  & 30.67 \\ 	
			4  &\Checkmark& (128, 32, 32, 128)   &  (256, 64, 64, 256)   &  1,488k  & 30.91 \\
			4  &\Checkmark& (256, 64, 64, 256)   &  (256, 64, 64, 256)   &   3,945k    & 30.96 \\				
			\bottomrule

		\end{tabular}

	\label{tab:ablation}
\end{table*}

\begin{table*}
	\centering
	\caption{
		Color image denoising results (PSNR) of different methods on the Kodak dataset. The best result is shown in bold.
	}
	\label{tab:psnr} 
	\resizebox{\textwidth}{!}{
		\begin{tabular}{|@{\hskip 0.2mm}c@{\hskip 0.2mm}||c|@{\hskip 0.2mm}c@{\hskip 0.2mm}|c|c|c||c|@{\hskip 0.2mm}c@{\hskip 0.2mm}|c|c|c||c|@{\hskip 0.2mm}c@{\hskip 0.2mm}|c|c|c|}
			\hline
			& \multicolumn{5}{c||}{$\sigma = 25$} & \multicolumn{5}{c||}{$\sigma = 35$} & \multicolumn{5}{c|}{$\sigma = 50$} \\
			\hline
			\hline
			Image & \textbf{CBM3D} & \textbf{MCWNNM} & \textbf{DnCNN} & \textbf{FFDNet} &  \textbf{Proposed}
			& \textbf{CBM3D} & \textbf{MCWNNM} & \textbf{DnCNN} & \textbf{FFDNet} &  \textbf{Proposed} 
			& \textbf{CBM3D} & \textbf{MCWNNM} & \textbf{DnCNN} & \textbf{FFDNet} &  \textbf{Proposed} \\
			\hline
			\hline				
			01 
            & 29.13 & 28.66 & 29.75 & 29.65 & \textbf{29.76}
			& 27.31 & 26.93 & 28.10 & 28.01 & \textbf{28.11}
			& 25.86 & 25.28 & 26.52 & 26.45 & \textbf{26.55} \\			
			\hline
			02 
            & 32.44 & 31.92 & 32.97 & 32.86 & \textbf{33.00}
			& 31.07 & 30.62 & 31.65 & 31.60 & \textbf{31.75}
			& 29.84 & 29.27 & 30.44 & 30.35 & \textbf{30.54} \\	
			\hline
			03 
            & 34.54 & 34.05 & 34.97 & 34.92 & \textbf{35.12}
			& 32.62 & 32.27 & 33.37 & 33.38 & \textbf{33.58}
			& 31.34 & 30.52 & 31.76 & 31.77 & \textbf{31.99} \\			
			\hline
			04 
            & 32.67 & 32.42 & 32.94 & 32.90 & \textbf{33.01}
			& 31.02 & 30.92 & 31.51 & 31.49 & \textbf{31.59}
			& 29.92 & 29.37 & 30.12 & 30.09 & \textbf{30.22} \\
			\hline
			05 
            & 29.73 & 29.37 & 30.53 & 30.35 & \textbf{30.55}
			& 27.61 & 27.53 & 28.66 & 28.52 & \textbf{28.72}
			& 25.92 & 25.60 & 26.77 & 26.67 & \textbf{26.87} \\			
			\hline
			06 
            & 30.59 & 30.18 & 31.05 & 30.99 & \textbf{31.08}
			& 28.78 & 28.44 & 29.37 & 29.34 & \textbf{29.45}
			& 27.34 & 26.70 & 27.74 & 27.72 & \textbf{27.85} \\	
			\hline
			07 
            & 33.66 & 33.36 & 34.42 & 34.31 & \textbf{34.47}
			& 31.64 & 31.53 & 32.60 & 32.54 & \textbf{32.70}
			& 29.99 & 29.51 & 30.67 & 30.66 & \textbf{30.82} \\			
			\hline
			08 
            & 29.88 & 29.39 & 30.30 & 30.04 & \textbf{30.37}
			& 27.82 & 27.67 & 28.53 & 28.30 & \textbf{28.64}
			& 26.23 & 25.86 & 26.65 & 26.50 & \textbf{26.84} \\
			\hline
			09 
            & 34.06 & 33.42 & 34.59 & 34.52 & \textbf{34.63}
			& 32.28 & 31.76 & 33.06 & 33.00 & \textbf{33.11}
			& 30.86 & 30.00 & 31.42 & 31.40 & \textbf{31.53} \\	
			\hline
			10 
            & 33.82 & 33.23 & 34.33 & 34.24 & \textbf{34.38}
			& 31.97 & 31.51 & 32.74 & 32.68 & \textbf{32.83}
			& 30.48 & 29.63 & 31.03 & 30.99 & \textbf{31.17} \\			
			\hline
			11 
            & 31.25 & 30.62 & 31.82 & 31.70 & \textbf{31.84}
			& 29.53 & 29.04 & 30.23 & 30.14 & \textbf{30.29}
			& 28.00 & 27.41 & 28.67 & 28.56 & \textbf{28.76} \\		
			\hline
			12 
            & 33.76 & 33.02 & 34.12 & 34.04 & \textbf{34.18}
			& 32.24 & 31.52 & 32.73 & 32.65 & \textbf{32.83}
			& 30.98 & 30.00 & 31.32 & 31.24 & \textbf{31.47} \\
			\hline					
			13 
            & 27.64 & 27.19 & \textbf{28.26} & 28.17 & 28.24
			& 25.70 & 25.40 & 26.46 & 26.39 & \textbf{26.47}
			& 24.03 & 23.70 & 24.73 & 24.67 & \textbf{24.76} \\			
			\hline
			14 
            & 30.03 & 29.67 & 30.79 & 30.67 & \textbf{30.80}
			& 28.24 & 28.05 & 29.17 & 29.07 & \textbf{29.20}
			& 26.74 & 26.43 & 27.57 & 27.48 & \textbf{27.63} \\	
			\hline
			15 
            & 33.08 & 32.69 & 33.32 & 33.24 & \textbf{33.35}
			& 31.47 & 31.15 & 31.89 & 31.82 & \textbf{31.96}
			& 30.32 & 29.59 & 30.50 & 30.44 & \textbf{30.59} \\			
			\hline
			16 
            & 32.33 & 31.79 & 32.69 & 32.66 & \textbf{32.74}
			& 30.64 & 30.15 & 31.16 & 31.14 & \textbf{31.23}
			& 29.36 & 28.53 & 29.68 & 29.63 & \textbf{29.78} \\
			\hline
			17 
            & 32.93 & 32.39 & \textbf{33.53} & 33.47 & 33.50
			& 30.64 & 30.75 & 31.96 & 31.94 & \textbf{31.98}
			& 29.36 & 28.98 & 30.33 & 30.33 & \textbf{30.40} \\			
			\hline
			18 
            & 29.83 & 29.46 & 30.40 & 30.31 & \textbf{30.46}
			& 28.00 & 27.70 & 28.72 & 28.63 & \textbf{28.79}
			& 26.41 & 25.94 & 27.03 & 26.96 & \textbf{27.14} \\	
			\hline
			19 
            & 31.78 & 31.29 & \textbf{32.23} & 32.12 & 32.30
			& 30.19 & 29.86 & 30.80 & 30.70 & \textbf{30.88}
			& 29.06 & 28.44 & 29.34 & 29.26 & \textbf{29.49} \\			
			\hline
			20 
            & 33.45 & 32.78 & 34.15 & 34.03 & \textbf{34.29}
			& 31.84 & 31.32 & 32.73 & 32.65 & \textbf{32.91}
			& 30.51 & 29.79 & 31.28 & 31.21 & \textbf{31.53} \\
			\hline
			21 
            & 30.99 & 30.55 & 31.61 & 31.53 & \textbf{31.63}
			& 29.17 & 28.86 & 29.94 & 29.88 & \textbf{29.98}
			& 27.61 & 27.13 & 28.27 & 28.21 & \textbf{28.34} \\	
			\hline
			22 
            & 30.93 & 30.48 & \textbf{31.41} & 31.33 & 31.38
			& 29.36 & 28.93 & 29.94 & 29.88 & \textbf{29.95}
			& 28.09 & 27.47 & 28.54 & 28.48 & \textbf{28.58} \\			
			\hline
			23 
            & 34.79 & 34.45 & 35.36 & 35.31 & \textbf{35.40}
			& 33.09 & 32.79 & 33.86 & 33.82 & \textbf{33.89}
			& 31.75 & 30.96 & 32.18 & 32.22 & \textbf{32.30} \\		
			\hline
			24 
            & 30.09 & 29.93 & \textbf{30.79} & 30.66 & 30.77
			& 28.19 & 28.17 & 28.98 & 28.89 & \textbf{29.03}
			& 26.62 & 26.37 & 27.18 & 27.11 & \textbf{27.30} \\
			\hline	
			\hline				
			\textbf{Average} 
            & 31.81 & 31.35 & 32.35 & 32.25& \textbf{32.39}
			& 30.04 & 29.70 & 30.76 & 30.69 & \textbf{30.83}
			& 28.62 & 28.02 & 29.16 & 29.10 & \textbf{29.27} \\																	
			\hline
		\end{tabular}
	}
\end{table*}

\subsection{Image Denoising}

Our proposed denoising network takes RGB images as input,  
and outputs the reconstructed images directly.
We add independent and identically distributed Gaussian noise with zero mean to the original image as the noisy input image during training.
We use the same training set as in~\cite{zhang2017beyond}, which is 432 color images from the Berkeley segmentation dataset \cite{martin2001database}.
The loss of separately training such a denoising network is equivalent to Equation~\ref{eqn:loss} as $\lambda_P = \lambda_H = 0$.
We use SGD with a batch size of 32, and the input patches are $48 \times 48$ pixels.
The initial learning rate is set as $10^{-4}$ and is divided by 10 after every 500,000 iterations. The training is terminated after 1,500,000 iterations.
We train a different denoising network for each noise level in our experiment.

\subsubsection{Analysis of Network Architecture}
\label{sec:ablation}

To investigate the architecture of our proposed denoising network, we design a set of controlled experiments with different network structures, 
in order to analyze a number of design choices   and examine their contributions to the overall denoising performance.
Specifically, we change the number of spatial scales of feature maps, the design of the encoding and decoding modules, as well as the existence of the global skip connection. 
Recall that we use one convolutional layer plus one residual block from~\cite{he2016deep} as the encoding and decoding modules, and we test the case in which the four convolutional layers in each module have (128, 32, 32, 128) kernels or (256, 64, 64, 256) kernels.   
We use these models to test on the widely used Kodak dataset,\footnote{\url{http://r0k.us/graphics/kodak/}} which consists of 24 color images with $\sigma = 35$.
The average PSNR (in dB) and the number of model parameters are displayed in Table ~\ref{tab:ablation}.
It is observed that the performance gets better when the computation cost becomes higher (i.e. more spatial scales and more model parameters). The global skip connection improves the results.
Considering the trade-off between the denoising performance and the computational cost, we adopt the encoding module of (128, 32, 32, 128) kernels and the decoding module of (256, 64, 64, 256) kernels with three spatial scales and the global skip connection in the rest of our experiments.




\begin{table}
	\begin{center}

		\caption{Color image denoising results (PSNR) of different methods on the CBSD68 dataset. The best result is shown in bold. }
		\label{tab:psnr_cbsd68}		
		\begin{tabular}{|c||c|c|c|c|c|}		
			\hline
			 & CBM3D & MCWNNM & DnCNN  & FFDNet & Proposed \\
			\hline
			\hline
			$\sigma = 25$ & 30.71 & 30.23 & 31.23 & 31.21 & \textbf{31.32} \\
			\hline
			$\sigma = 35$ & 28.89 & 28.57 & 29.58 & 29.58 & \textbf{29.73} \\
			\hline
			$\sigma = 50$ & 27.38 & 26.63 & 27.92 & 27.76 & \textbf{28.09} \\
			\hline	
		\end{tabular}
		
	\end{center}
\end{table} 


		

\begin{figure*}[!th]
	\center
	\resizebox{\textwidth}{!}{
	\begin{tabular}{c@{\hskip 1mm}c@{\hskip 1mm}c@{\hskip 1mm}c@{\hskip 1mm}c}

		\includegraphics[width=0.19\linewidth]{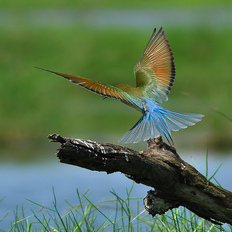} &
		\includegraphics[width=0.19\linewidth]{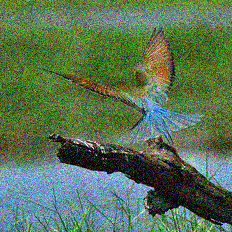} &
		\includegraphics[width=0.19\linewidth]{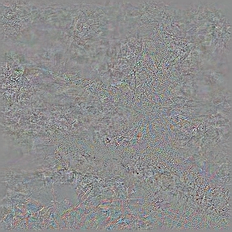} &
		\includegraphics[width=0.19\linewidth]{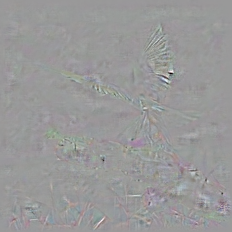} &
		\includegraphics[width=0.19\linewidth]{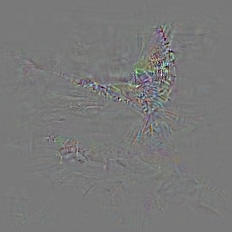} \\
		
		\includegraphics[width=0.19\linewidth]{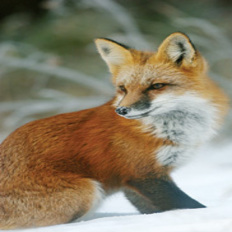} &
		\includegraphics[width=0.19\linewidth]{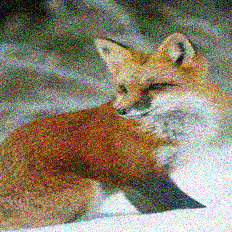} &
		\includegraphics[width=0.19\linewidth]{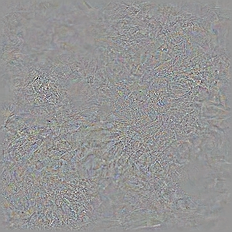} &
		\includegraphics[width=0.19\linewidth]{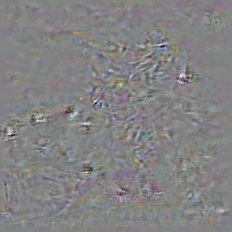} &
		\includegraphics[width=0.19\linewidth]{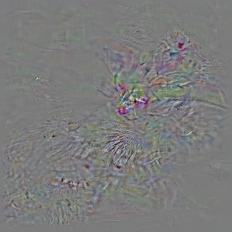} \\
		
		\includegraphics[width=0.19\linewidth]{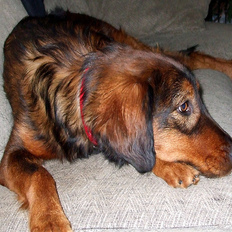} &
		\includegraphics[width=0.19\linewidth]{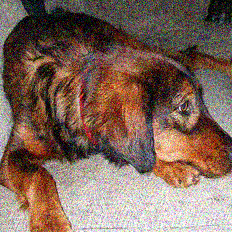} &
		\includegraphics[width=0.19\linewidth]{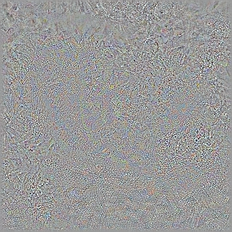} &
		\includegraphics[width=0.19\linewidth]{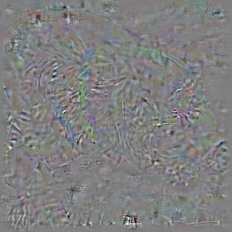} &
		\includegraphics[width=0.19\linewidth]{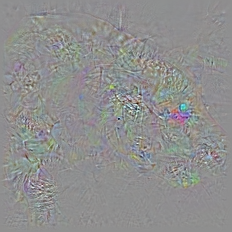} \\
		
		\includegraphics[width=0.19\linewidth]{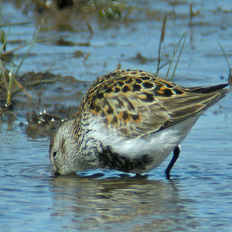} &
		\includegraphics[width=0.19\linewidth]{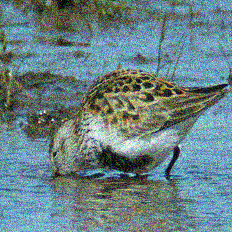} &
		\includegraphics[width=0.19\linewidth]{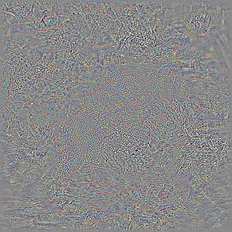} &
		\includegraphics[width=0.19\linewidth]{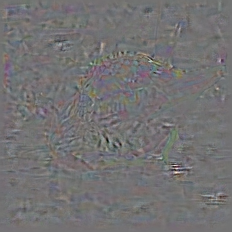} &
		\includegraphics[width=0.19\linewidth]{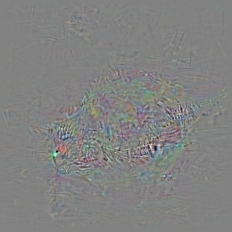} \\
		
		(a)  & (b)  & (c)  & (d)  & (e) \\
		
	\end{tabular}
	}
	\caption{From left to right: (a) Ground truth images in ILSVRC2012 validation set. (b) Noisy images. (c) The visualization of gradient with respect to the input noisy image for only the reconstruction loss. (d) The visualization of gradient with respect to the input noisy image for only the perceptual loss. (e) The visualization of gradient with respect to the input noisy image for only the classification loss.}
	\label{fig:viz}
\end{figure*}

\subsubsection{Comparisons with State-of-the-Art Methods}

We compare our denoising network with several state-of-the-art color image denoising approaches on various noise levels: $\sigma = 25, 35$ and $50$.
We evaluate their denoising performance
over the following two datasets: the Kodak dataset,
and the CBSD68 dataset which consists of 68 color images as in \cite{zhang2018ffdnet}.
Table~\ref{tab:psnr} and~\ref{tab:psnr_cbsd68} shows the peak signal-to-noise ratio (PSNR) results for CBM3D \cite{dabov2007color},  MCWNNM~\cite{xu2017multi}, DnCNN~\cite{zhang2017beyond},  FFDNet~\cite{zhang2018ffdnet} and our proposed method
on these two datasets, respectively. 
It is noteworthy that we calculate PSNRs on the whole images of the Kodak dataset, rather than the cropped regions used in \cite{zhang2018ffdnet}.
We do not list other methods \cite{burger2012image,zoran2011learning,gu2014weighted,chen2017trainable,zhang2017learning} whose performance is worse than DnCNN or FFDNet.
The implementation codes used are from the authors' websites and the default parameter settings are adopted in our experiments.


It is clear that our proposed method outperforms all the competing approaches quantitatively across different noise levels.
It achieves the highest PSNR in almost every image of the Kodak dataset, and obtains the highest average PSNR over the CBSD68 dataset.



\begin{figure*}[!th]
	\center
	\begin{tabular}{@{\hskip 0.5mm}c@{\hskip 0.5mm}c@{\hskip 1mm}c@{\hskip 0.5mm}c}
		\multicolumn{2}{c}{\includegraphics[height=0.19\linewidth, trim=0 0 2 309,clip]{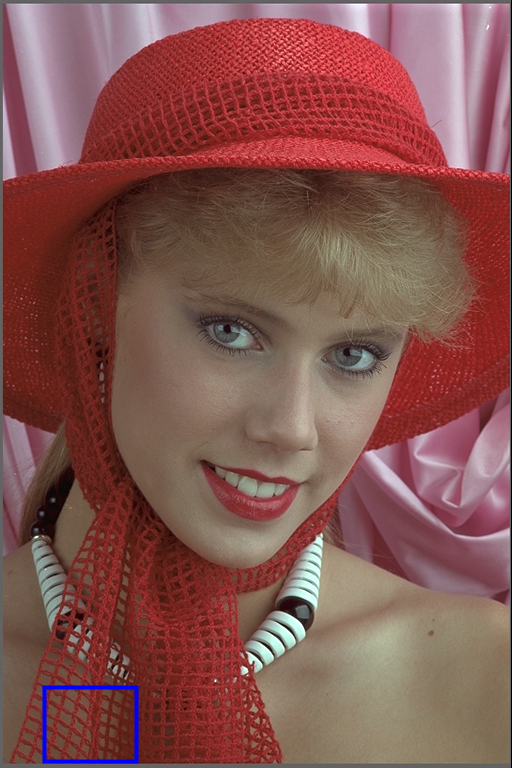}} &
		\multicolumn{2}{c}{\includegraphics[height=0.19\linewidth, trim=88 0 40 32,clip]{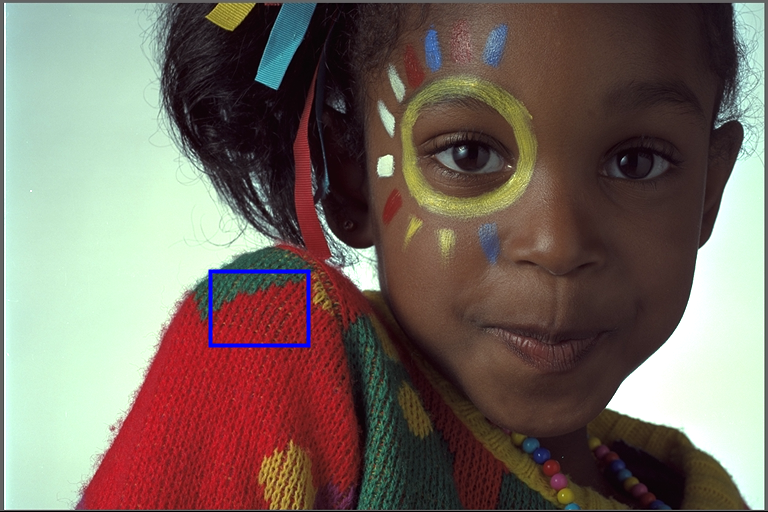}} \\
		\multicolumn{2}{c}{(I)} &
		\multicolumn{2}{c}{(I)} \\		
		\includegraphics[height=0.19\linewidth]{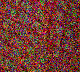} &
		\includegraphics[height=0.19\linewidth]{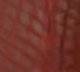} &
		\includegraphics[height=0.19\linewidth]{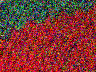} &
		\includegraphics[height=0.19\linewidth]{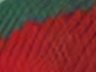}  \\
		 (II)  & (III)  & (II)  & (III)  \\
		\includegraphics[height=0.19\linewidth]{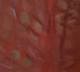} &
		\includegraphics[height=0.19\linewidth]{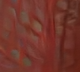} &		
		\includegraphics[height=0.19\linewidth]{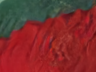} &
		\includegraphics[height=0.19\linewidth]{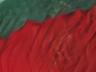} \\
		(IV)  & (V)  & (IV)  & (V) \\	
		\includegraphics[height=0.19\linewidth]{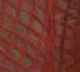}	&
		\includegraphics[height=0.19\linewidth]{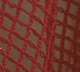} &		
		\includegraphics[height=0.19\linewidth]{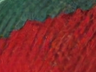}  &
		\includegraphics[height=0.19\linewidth]{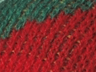} \\
		(VI)  & (VII) &  (VI)  & (VII) \\
		\multicolumn{2}{c}{(a)} & \multicolumn{2}{c}{(b)} \\
	\end{tabular}
	\caption{Two image denoising examples from the Kodak dataset with noise level of 50 are displayed in (a) and (b). We show (I) the ground truth image and the zoom-in regions of: (II) the noisy image; (III) the denoised image by CBM3D; (IV) the denoised image by DnCNN; the denoising result of our proposed model (V) without the guidance of high-level vision information; (VI) with the guidance of high-level vision information and (VII) the ground truth. }
	\label{fig:denoise3}
\end{figure*}

\subsection{When Image Denoising Meets High-Level Vision Tasks}

We choose two high-level vision tasks as representatives in our study: image classification and semantic segmentation \cite{wang2018understanding}, which have been dominated by deep network based models. 
We utilize two popular VGG-based deep networks in our system for each task, respectively.
\textit{VGG-16} in~\cite{simonyan2014very} is employed for image classification;
we select \textit{DeepLab-LargeFOV} in~\cite{chen2014semantic} for semantic segmentation.
We follow the preprocessing protocols (e.g. crop size, mean removal of each color channel) in~\cite{simonyan2014very} and~\cite{chen2014semantic} accordingly while training and deploying them in our experiments.

As for the cascaded network for image classification and the corresponding experiments, we train our model on ILSVRC2012 training set, 
and evaluate the classification accuracy on ILSVRC2012 validation set. 
$\lambda_P$ and $\lambda_H$ are empirically set as $0.5$ and $0.25$, respectively.
As for the cascaded network for image semantic segmentation and its corresponding experiments, we train our model on the augmented training set of Pascal VOC 2012 as in~\cite{chen2014semantic}, and test on its validation set. 
$\lambda_P$ and $\lambda_H$ are empirically set as $0.5$ and $0.5$, respectively.

\subsubsection{High-Level Vision Information Guided Image Denoising}

The typical metric used for image denoising is PSNR, which has been shown to 
sometimes correlate poorly with human assessment of visual quality~\cite{huynh2008scope}. 
Since PSNR depends on the reconstruction error between the denoised output and the reference image, a model trained by minimizing MSE on the image domain should always outperform a model trained by minimizing our proposed joint loss (with the guidance of high-level vision semantics) in the metric of PSNR.
Therefore, we emphasize that the goal of our following experiments is not to pursue the highest PSNR, but to demonstrate the qualitative difference between the model trained with our proposed joint loss and the model trained with MSE on the image domain.

In our cascade architecture, we study the mutual influence between image denoising and high-level vision tasks. 
Considering image classification as an example of high-level vision tasks,
the loss used in our framework is composed of (1) the image reconstruction loss from the denoising network, (2) the perceptual loss from the VGG-based network, and (3) the classification loss from the VGG-based network.
First we analyze the contribution of each loss component, by visualizing the gradient with respect to the input noisy image for applying only (1) the image reconstruction loss, (2) the perceptual loss, and (3) the classification loss, respectively.
Several examples from ILSVRC2012 validation set are displayed in Figure~\ref{fig:viz}.

Figure~\ref{fig:viz}(c) shows that the energy is spread out across the visualization of gradient map with respect to the input noisy image for only the reconstruction loss, 
since the additive noise is distributed ubiquitously in the same level over the image, and the reconstruction loss is not spatially biased.
In contrast,  Figure~\ref{fig:viz}(e) shows the visualization of gradient map with respect to the input noisy image for only the classification loss, where the energy is more concentrated on regions with class-specific details, which usually contain key features to recognize the class of images.
These regions are generally the texture-rich regions that are prone to suffer from oversmoothing in conventional denoising approaches, 
such as the fur of fox and dog, as well as the feather of bird in Figure~\ref{fig:viz}.
Figure~\ref{fig:viz}(d) is the visualization of gradient map with respect to the input noisy image for only the perceptual loss, and the energy tends to focus on a number of local features.
In other words, with the class information as auxiliary supervision, our denoising network trained with the joint loss is able to denoise such class-specific regions differently from the rest of image, 
which demonstrates that
these three types of losses
work in a complementary manner for denoising.

Figure~\ref{fig:denoise3} displays two image denoising examples from the Kodak dataset. 
A visual comparison is illustrated for a zoom-in region: (II) and (III) are the denoising results using CBM3D~\cite{dabov2007color}, and DnCNN~\cite{zhang2017beyond}, respectively; (IV) is the proposed denoiser trained separately without the guidance of high-level vision information; (V) is the denoising result using the proposed denoising network trained jointly with a segmentation network.
We can find that the results using CBM3D, DnCNN and our separately trained denoiser generate oversmoothing regions, while the jointly trained denoising network is able to reconstruct the denoised image which preserves more details and textures with better visual quality. 

\begin{figure}[th]
	\centering
	\includegraphics[width=0.9\linewidth]{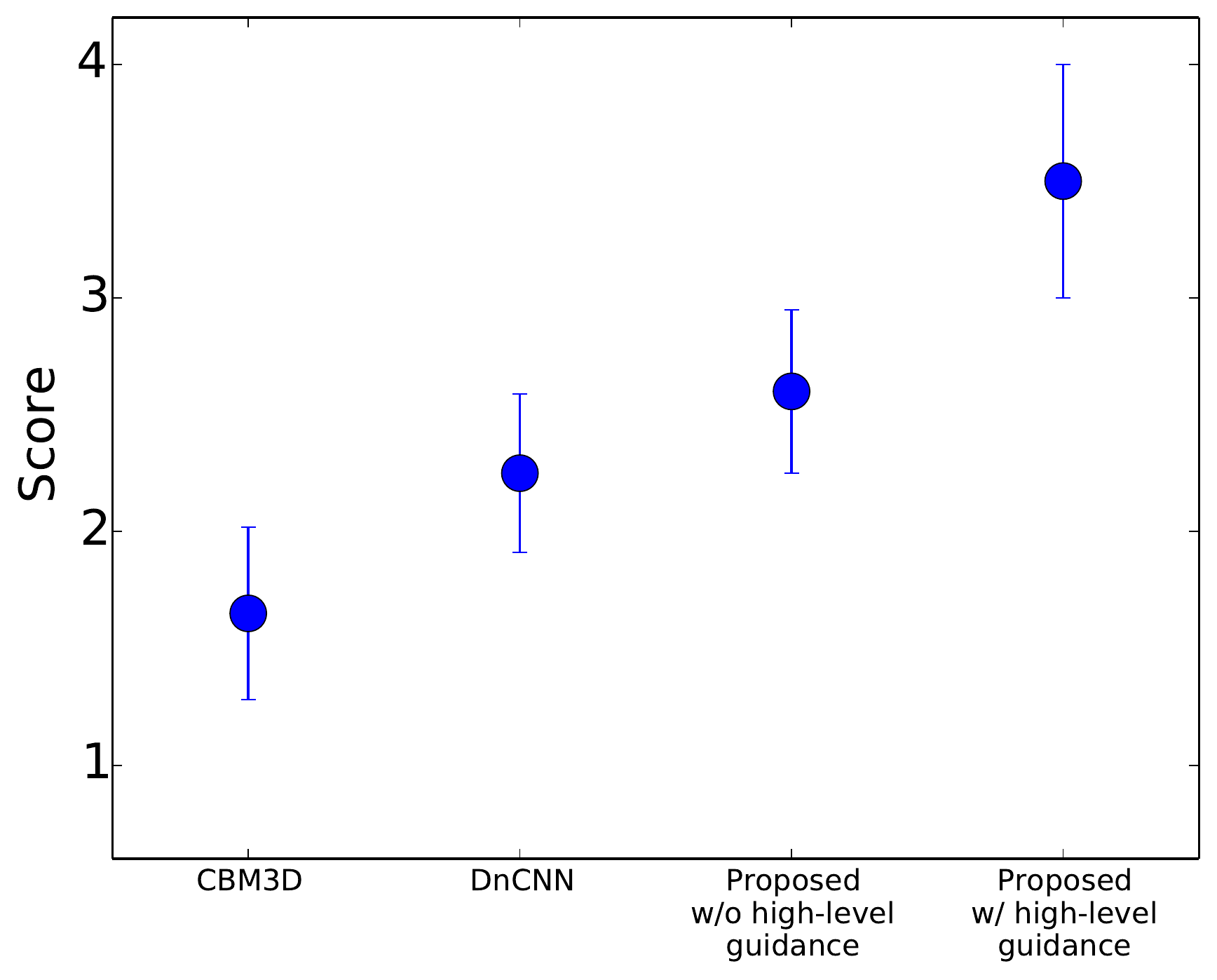}
	\caption{Mean opinion scores of the subjective evaluation for four denoising methods. The longer bar indicates the larger variance of the scores.}
	\label{fig:MOS}
\end{figure}

In order to more thoroughly quantify the perceptual quality of the denoised results, we perform a mean
opinion score (MOS) test. 
We compare the following four denoising methods: CBM3D, DnCNN, our proposed denoising network without high-level guidance and with high-level guidance.
In our test, for each noisy image the denoised results from these four methods are shown simultaneously. 
We use scores from 1 to 4 to indicate from low image quality to high image quality, and ask the raters to rank the denoised results by assigning corresponding scores. 
15 raters are invited to participate our test.
In each test, 5 images from the Kodak dataset and 5 images from the CBSD68 dataset are randomly selected and the denoised results are from the noise level of 50. 
The MOS result of our perceptual evaluation is shown in Figure~\ref{fig:MOS}.
We find that our proposed denoising method with high-level guidance achieves the highest score, showing high-level semantics is able to improve the visual perception in our proposed framework.

\subsubsection{Generality of the Denoiser for High-Level Vision Tasks}
\label{exp_high_level}

\begin{figure*}[t!]
\begin{center}
\begin{tabular}{c@{\hskip 0.5mm}c@{\hskip 0.5mm}c@{\hskip 0.5mm}c}
\includegraphics[width=0.24\linewidth, trim=0 0 0 36, clip]{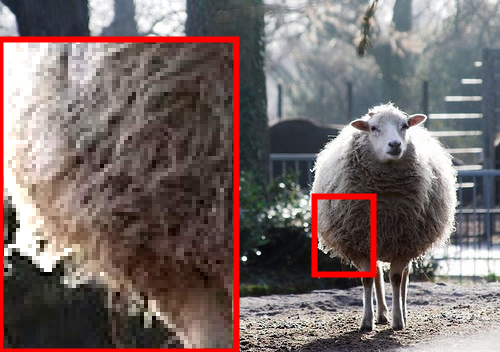} &
\includegraphics[width=0.24\linewidth, trim=0 0 0 36, clip]{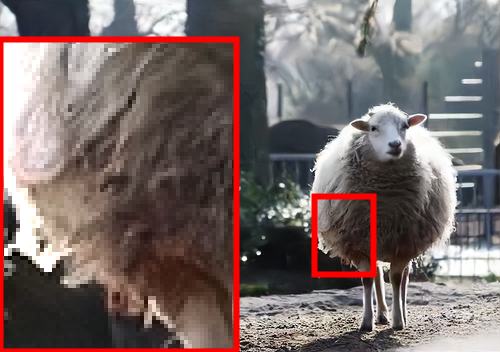} &
\includegraphics[width=0.24\linewidth, trim=0 0 0 36, clip]{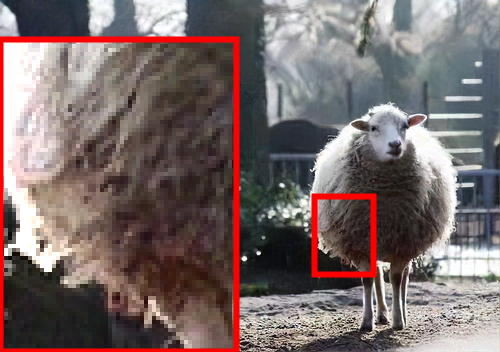} & 
\includegraphics[width=0.24\linewidth, trim=0 0 0 36, clip]{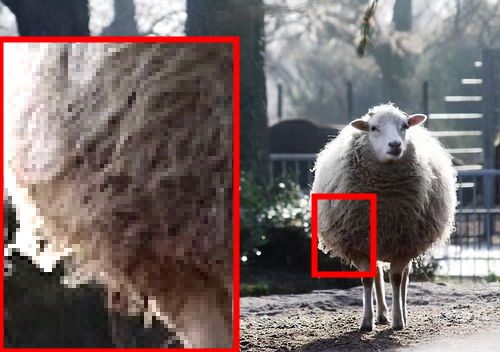} \\
\includegraphics[width=0.24\linewidth, trim=0 0 0 36, clip]{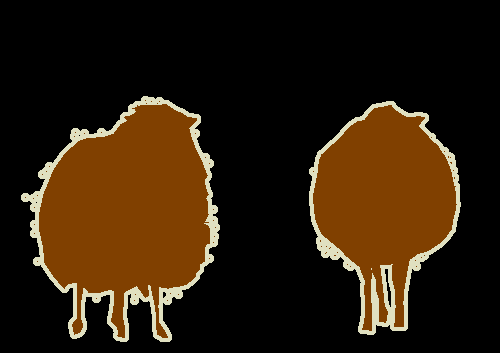} &
\includegraphics[width=0.24\linewidth, trim=0 0 0 36, clip]{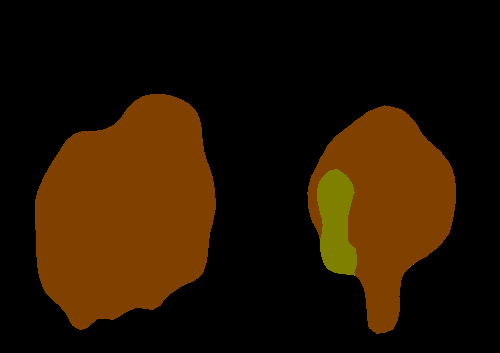} &
\includegraphics[width=0.24\linewidth, trim=0 0 0 36, clip]{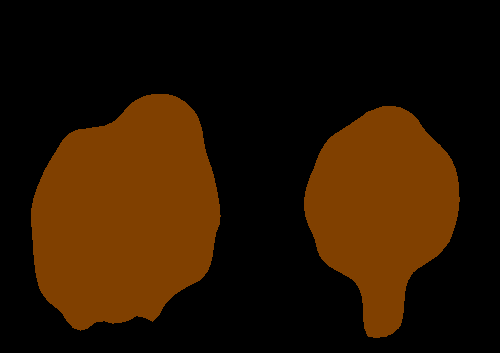} & 
\includegraphics[width=0.24\linewidth, trim=0 0 0 36, clip]{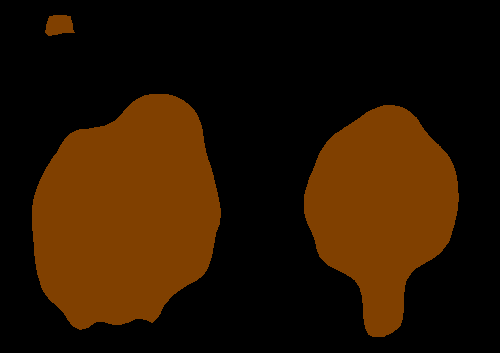} \\

\includegraphics[width=0.24\linewidth]{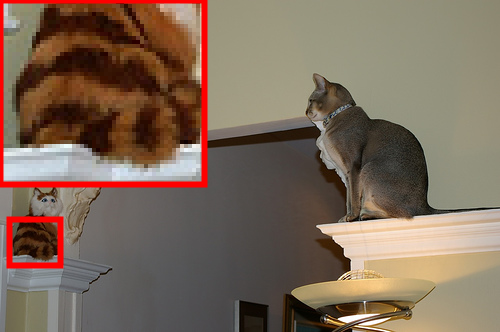} &
\includegraphics[width=0.24\linewidth]{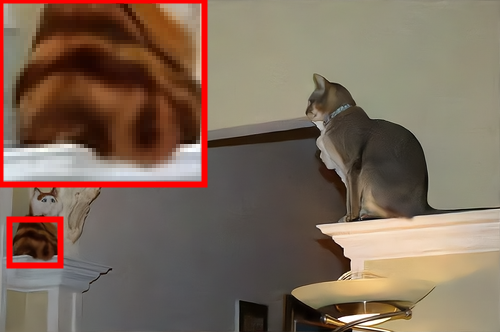} &
\includegraphics[width=0.24\linewidth]{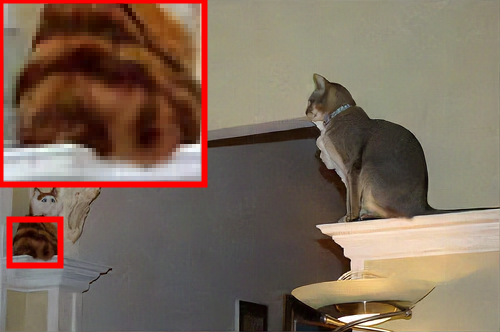} & 
\includegraphics[width=0.24\linewidth]{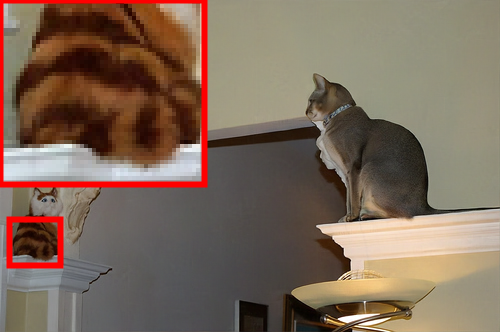} \\
\includegraphics[width=0.24\linewidth]{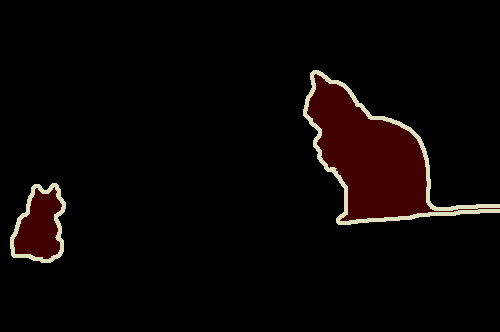} &
\includegraphics[width=0.24\linewidth]{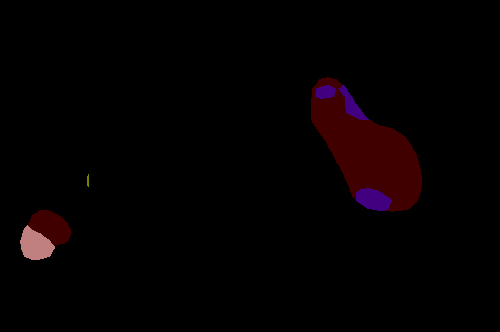} &
\includegraphics[width=0.24\linewidth]{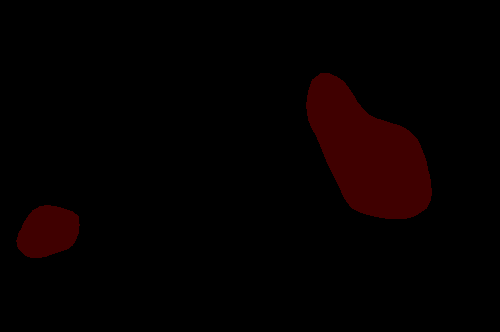} & 
\includegraphics[width=0.24\linewidth]{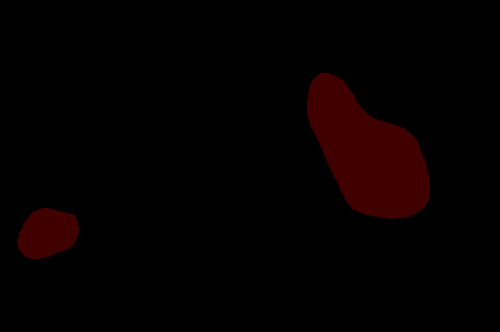} \\
{\small (a)} & {\small (b)} & {\small (c)} & {\small (d)} \\
\end{tabular}
\end{center}
\caption{Two semantic segmentation examples from Pascal VOC 2012 validation set. From left to right: (a) the ground truth image, the denoised image using (b) the separately trained denoiser, (c) the denoiser trained with the reconstruction and segmentation joint loss, and (d) the denoiser trained with the classification network and evaluated for semantic segmentation. 
Their corresponding segmentation label maps are shown below. The zoom-in region which generates inaccurate segmentation in (b) is displayed in the red box.
}
\label{fig:seg1}
\end{figure*}

We now investigate how the image denoising can enhance the high-level vision applications, including image classification and semantic segmentation, over the ILSVRC2012 and Pascal VOC 2012 datasets, respectively.
The noisy images ($\sigma = 15, 30, 45, 60$) are denoised and then fed into the VGG-based networks for high-level vision tasks. 
To evaluate how different denoising schemes contribute to the  performance of high-level vision tasks, we experiment with the following cases:

\begin{table}
	\begin{center}

		\caption{Classification accuracy after denoising noisy image input, averaged over ILSVRC2012 validation dataset. \textcolor{red}{Red} is the best and \textcolor{blue}{blue} is the second best results. }
		\label{table:classDenoising1}		
		\begin{tabular}{|@{\hskip 1mm}c@{\hskip 1mm}|@{\hskip 1mm}c@{\hskip 1mm}|c|@{\hskip 1mm}c@{\hskip 1mm}|@{\hskip 1mm}c@{\hskip 1mm}|@{\hskip 1mm}c@{\hskip 1mm}|@{\hskip 1mm}c@{\hskip 1mm}|}		
			\hline
			\multicolumn{2}{|c|}{} & VGG & CBM3D + & Separate + & Joint & Joint Training\\   [-0.3ex] 
			\multicolumn{2}{|c|}{} & & VGG & VGG & Training & (Cross-Task) \\   
			\hline 
			\hline
			\multirow{2}{*}{$\sigma$=15} 
			& Top-1 & 62.4 & 68.2 & 68.3 & \textcolor{red}{69.9} & \textcolor{blue}{69.8} \\ [-0.2ex]
			& Top-5 & 84.2 & 88.8 & 88.7 & \textcolor{red}{89.5} & \textcolor{blue}{89.4} \\ 
			\hline
			\multirow{2}{*}{$\sigma$=30} 
			& Top-1 & 44.4 & 62.3 & 62.7 & \textcolor{red}{67.1} & \textcolor{blue}{66.4} \\ [-0.2ex]
			& Top-5 & 68.9 & 84.8 & 84.9 & \textcolor{red}{87.6} & \textcolor{blue}{87.2} \\ 
			\hline
			\multirow{2}{*}{$\sigma$=45} 
			& Top-1 & 24.3 & 55.2 & 54.6 & \textcolor{red}{63.2} & \textcolor{blue}{62.1} \\ [-0.2ex]
			& Top-5 & 46.1 & 79.4 & 78.8 & \textcolor{red}{84.7} & \textcolor{blue}{84.0} \\ 
			\hline
			\multirow{2}{*}{$\sigma$=60} 
			& Top-1 & 11.4 & 50.0 & 50.1 & \textcolor{red}{59.4} & \textcolor{blue}{57.2} \\ [-0.2ex]
			& Top-5 & 26.3 & 74.2 & 74.5 & \textcolor{red}{81.9} & \textcolor{blue}{80.3} \\ 
			\hline	
		\end{tabular}
		
	\end{center}
\end{table}

\begin{itemize}
	\item Noisy images are directly fed into the high-level vision network, termed as \textit{VGG}. This approach serves as the baseline.
	\item Noisy images are first denoised by CBM3D, and then fed into the high-level vision network, termed as \textit{CBM3D+VGG}.
	\item Noisy images are denoised via the separately trained denoising network, and then fed into the high-level vision network, termed as \textit{Separate+VGG}.
	\item Our proposed approach: noisy images are processed by the cascade of these two networks, which is trained using the joint loss, termed as \textit{Joint Training}.
	\item A denoising network is trained with the classification network in our proposed approach, but then is connected to the segmentation network and evaluated for the task of semantic segmentation, or vice versa. This is to validate the generality of our denoiser for various high-level tasks, termed as \textit{Joint Training (Cross-Task).
}
\end{itemize}
Note that the weights in the high-level vision network are initialized from a well-trained network under the noiseless setting and not updated during training in our experiments.


\begin{table}
	\begin{center}
		\caption{Segmentation results (mIoU) after denoising noisy image input, averaged over  Pascal VOC 2012 validation dataset. \textcolor{red}{Red} is the best and \textcolor{blue}{blue} is the second best results.}	
		\label{table:classDenoising2}				
		\begin{tabular}{|c|c|@{\hskip 1mm}c@{\hskip 1mm}|@{\hskip 1mm}c@{\hskip 1mm}|@{\hskip 1mm}c@{\hskip 1mm}|@{\hskip 1mm}c@{\hskip 1mm}|}		
			\hline
			 & VGG & CBM3D + & Separate + & Joint & Joint Training \\   [-0.3ex] 
			 & & VGG & VGG &  Training & (Cross-Task) \\   
			\hline 
			\hline
			
			$\sigma$=15 & 56.78 & 59.58 & 58.70 & \textcolor{red}{60.47} & \textcolor{blue}{60.41} \\
			\hline	
			
			$\sigma$=30  & 43.43 & 55.29 & 54.13 & \textcolor{red}{57.88} & \textcolor{blue}{56.30} \\
			\hline
			
			$\sigma$=45  & 27.99 & 50.69 & 49.51 & \textcolor{red}{54.84} & \textcolor{blue}{54.03} \\
			\hline
			
			$\sigma$=60  & 14.94 & 46.56 & 46.59 & \textcolor{red}{52.05} & \textcolor{blue}{51.84} \\
			\hline			
		\end{tabular}
		
	\end{center}
\end{table}

Table~\ref{table:classDenoising1} and Table~\ref{table:classDenoising2} list the performance of high-level vision tasks, i.e., top-1 and top5 accuracy for classification and mean intersection-over-union (IoU) without conditional random field (CRF) postprocessing for semantic segmentation. 
We notice that the baseline VGG approach obtains much lower accuracy than all the other cases, which shows the necessity of image denoising as a preprocessing step for high-level vision tasks on noisy data.
When we only apply denoising without considering high-level semantics (e.g., in CBM3D+VGG and Separate+VGG), it also fails to achieve high accuracy due to the artifacts introduced by the denoisers.
The proposed Joint Training approach achieves sufficiently high accuracy across various noise levels.


As for the case of Joint Training (Cross-Task), first we train the denoising network jointly with the segmentation network and then connect this denoiser to the classification network.
As shown in Table~\ref{table:classDenoising1},  
its accuracy remarkably outperforms the cascade of a separately trained denoising network and a classification network (i.e., Separate+VGG), and is comparable to our proposed model dedicatedly trained for classification (Joint Training). 
In addition, we use the denoising network jointly trained with the classification network to connect the segmentation network. Its mean IoU is much better than Separate+VGG in Table~\ref{table:classDenoising2}. 
These two experiments show that the high-level semantics of different tasks are universal in terms of low-level vision tasks, which is in line with intuition, and the denoiser trained in our method has the generality for various high-level tasks.

Figure~\ref{fig:seg1} displays two visual examples of how the data-driven denoising can enhance the semantic segmentation performance. 
It is observed that the segmentation result of the denoised image from the separately trained denoising network has lower accuracy compared to those using the joint loss and the joint loss (cross-task), while the zoom-in region of its denoised image for inaccurate segmentation in Figure~\ref{fig:seg1}~(b) contains oversmoothing artifacts.
In contrast, both the Joint Training and Joint Training (Cross-Task) approaches achieve finer segmentation result and produce more visually pleasing denoised outputs simultaneously. 

\subsection{Extension to Real Noise Removal}

\begin{figure}[t]
    \resizebox{\columnwidth}{!}{
		\begin{tabular}{@{\hskip 0.5mm}c@{\hskip 1mm}c@{\hskip 1mm}c@{\hskip 1mm}c@{\hskip 0.5mm}}
			\includegraphics[width=0.32\linewidth]{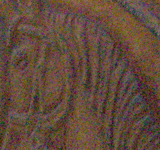} &
			\includegraphics[width=0.32\linewidth]{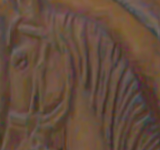} &
			\includegraphics[width=0.32\linewidth]{0026/zoom_BM3D_scaled.png} &
			\includegraphics[width=0.32\linewidth]{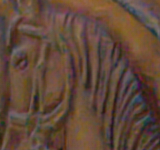} \\			
			\includegraphics[width=0.32\linewidth]{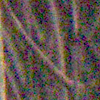} &
			\includegraphics[width=0.32\linewidth]{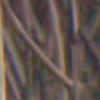} &					\includegraphics[width=0.32\linewidth]{0023/zoom_BM3D_scaled.png} &

			\includegraphics[width=0.32\linewidth]{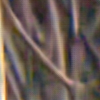} \\
			
			(a) & (b) & (c) & (d) \\
		\end{tabular}
    }
	\caption{Two denoising examples on real noisy images from DND. (a) the crop region of original images. (b) the denoising results from CBM3D. (c) the denoising results from DnCNN. (d) the denoising results from our method, i.e., the denoiser trained with the joint loss.
	}
	\label{fig:real1}
\end{figure}

Synthetic Gaussian noise removal has been extensively studied in previous works~\cite{Aharon2006,burger2012image,dabov2007image,wen2015octobos}. However, it is unclear if denoisers can be generalized from removing synthetic noise to handling realistic noise. Therefore, we
further test our method on real noisy images in the Darmstadt Noise Dataset (DND)~\cite{plotz2017benchmarking}, 
where the images are taken with higher ISO, and the corruption can be more complicated than i.i.d.~Gaussian noise~\cite{liu2008automatic}.
Denoising examples using CBM3D, DnCNN and our proposed method are shown in Figure~\ref{fig:real1}. 
It can be seen that CBM3D and DnCNN are prone to generate oversmoothing regions, while our proposed method preserves more local edges and high-frequency components, which leads to denoised images with better visual quality.
This shows that our denoising method is robust to the noise statistical distribution, and thus demonstrates good generalization to realistic corruption.

\section{Conclusion}
\label{sec:conc}

Investigating the relation between low-level image processing and high-level semantic tasks has great practical value in various applications of computer vision.
In this paper, we handle these two components in a simple yet efficient way by allowing the high-level semantic information to flow back to the low-level image processing part,
which achieves superior performance in both image denoising and various high-level vision tasks.
In addition, the denoiser trained for one high-level vision task in this manner has the robustness to other high-level vision tasks.
Overall, it provides a feasible and robust solution in a deep learning fashion to real world problems,
For future work, we will explore to embed the high-level semantics to more low-level vision tasks, e.g., super-resolution~\cite{liu2016robust,liu2018learning}, as well as to bring more types of semantical information into consideration.

\section*{Acknowledgments}
This work was supported by the U.S. Army Research Office under Grant W911NF-15-1-0317.

\bibliographystyle{IEEEtran}
\bibliography{ref}

\end{document}